# What Is Working Memory and Mental Imagery?
# A Robot that Learns to Perform Mental Computations

### Victor Eliashberg

*Avel Electronics, Palo Alto, California*
*October 2002, www.brain0.com*

> Turing's "Machines". These machines are humans who calculate.
>
> Ludwig Wittgenstein

## 1.0 Introduction

This paper goes back to Turing (1936) and treats his machine as a cognitive model **(W,D,B)**, where **W** is an "external world" represented by memory device (the tape divided into squares), and **(D,B)** is a simple robot that consists of the sensory-motor devices, D, and the brain, B. The robot's sensory-motor devices (the "eye", the "hand", and the "organ of speech") allow the robot to simulate the work of any Turing machine. The robot simulates the internal states of a Turing machine by "talking to itself."

At the stage of training, the teacher forces the robot (by acting directly on its motor centers) to perform several examples of an algorithm with different input data presented on tape. Two effects are achieved: 1) the robot learns to perform the shown algorithm with any input data using the tape; 2) the robot learns to perform the algorithm "mentally" using an "imaginary tape."

The model illustrates the simplest concept of a universal learning neurocomputer, demonstrates universality of associative learning as the mechanism of programming, and provides a simplified, but nontrivial neurobiologically plausible explanation of the phenomena of working memory and mental imagery. The model is implemented as a user-friendly program for Windows called EROBOT. The program is available at www.brain0.com/software.html. (This pre-print server doesn't allow one to go directly to external links, so you have to copy this URL).

A detailed theory of the discussed model was first described in Eliashberg (1979). It was later shown (Eliashberg 1989, 1990, and 2003) how the dynamics of the working memory of this model could be connected with the statistical dynamics of conformations of ion channels.

The paper includes the following sections:

**Section 1.1 Turing's Machine as a System (Robot,World)** This section goes back to Turing (1936) and treats his machine as a cognitive model of system (Man,World).

**Section 1.2 System-Theoretical Background** introduces some basic system-theoretical concepts and notation needed for understanding this paper.

**Section 1.3 Neurocomputing Background** provides some neurocomputing background needed for understanding the neural model discussed in Section 1.4.

**Section 1.4 Designing a Neural Brain for Turing's Robot** describes an associative neural network that can serve as the brain for the robot of Section 1.1.

**Section 1.5 Associative Neural Networks as Programmable Look-up Tables** illustrates two levels of formalism by replacing the "neurobiological" model of Section 1.4, expressed in terms of differential equations, with a more understandable "psychological" model expressed in terms of "elementary procedures."

**Section 1.6 Robot That Learns to Perform Mental Computations** enhances the sensory-motor devices, D, and the brain, B, of the robot described in Section 1.1. This enhancement gives the robot "working memory" and allows it to learn to perform "mental" computations with the use of an "imaginary" memory device.

**Section 1.7 Experiments with EROBOT** discusses some educational experiments with the program EROBOT that simulates the robot from Section 1.6. This section also serves as the user's manual describing the user interface of the program. As mentioned above, the program is available at www.brain0.com/software.html.

**Section 1.8 Basic Questions** discusses several basic questions related to the brain that are addressed by EROBOT.

**Section 1.9 Whither from Here** outlines some possibilities for the further development of the basic ideas illustrated by EROBOT.

**REFERENCES**

## 1.1 Turing's Machine as a System (Robot,World)

Consider the cognitive system (W,D,B) shown in Figure 1.1. The diagram illustrates my vision of the idea of Turing's machine in its original biological interpretation (Turing, 1936). The reader not familiar with the concept of a Turing machine should read the part of Turing's original paper that describes the way of thinking that led him to the invention of his machine. A good description of Turing's ideas can be found in Minsky (1967). It is interesting to mention that Turing used the term "computer" to refer to a person performing computations.

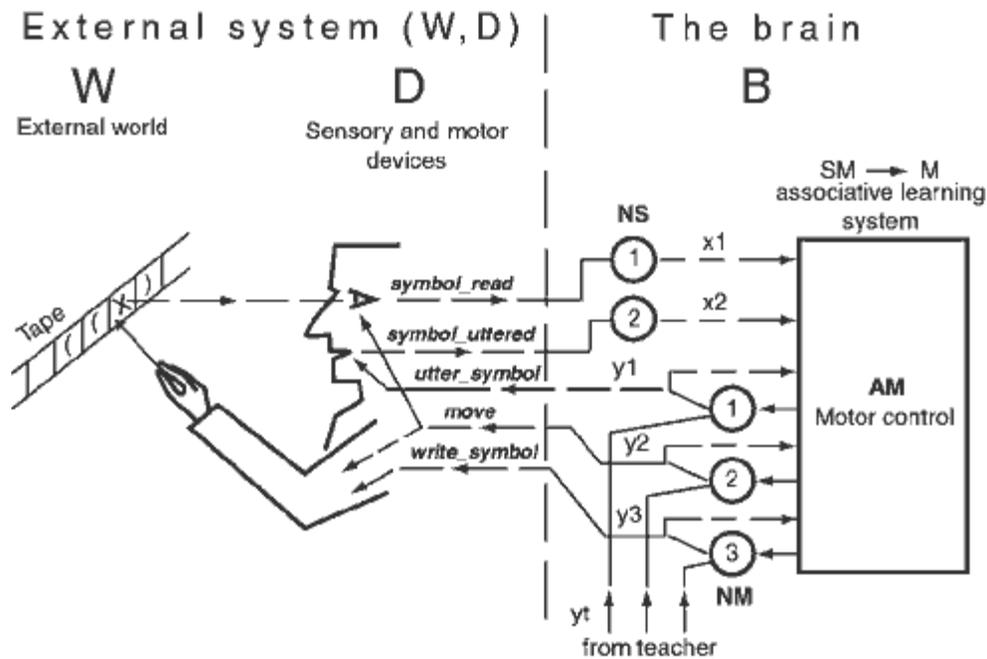

Figure 1.1 Turing machine as a system (Robot,World)

To be able to simulate the work of an arbitrary Turing machine the robot (system (D,B) ) shown in Figure 1.1 needs to perform the following elementary operations:

> Read a symbol from a single square scanned by the robot's eye. This square is called the *scanned square*.
>
> 1. Write a symbol into the scanned square. The previous symbol in the square is replaced by the new one.
>
> 2. Move the eye and the hand simultaneously to the new square called the *next square*. It is sufficient to be able to move one square to the left, one square to the right, or stay in the same square. (It doesn't hurt if the robot can move more, but it is not necessary.)

3. "Utter" a symbol representing the robots "intensions" for the next step of computations. (It is sufficient to keep this symbol "in mind" for just one time step.)

**Note.** In the model of Figure 1.1 this one-step memory is provided by the delayed feedback between the motor signal *utter_symbol* and the proprioceptive image of this motor symbol represented by the signal *symbol_uttered*.

**Teaching.** To teach the robot, the teacher acts directly on the robot's motor centers, NM. The teacher forces the robot to work as a Turing machine with several sample input data presented on tape. The goal of system AM is to learn to simulate the work of the teacher with any input data.

**Examination.** The teacher presents new input data written on tape. To pass the exam the robot has to correctly perform the demonstrated algorithm without the teacher.

**Note.** In a more complex cognitive model discussed in Section 1.6 (Figure 1.16) the robot will be required to perform the demonstrated algorithm without seeing the tape.

**Problem of synthesis.** Our first goal is to design associative learning system AM providing the described performance of the robot of Figure 1.1. We want to make our model neurobiologically consistent, so we shall try to use only those computational resources which can be reasonably postulated in biological neural networks.

## 1.2 System-Theoretical Background

Before proceeding with the problem of synthesis formulated in the previous section I need to define some basic system-theoretical concepts and notation needed for dealing with this problem. The reader who is familiar with these concepts should still read this section to make sure that we are using the same definitions.

### 1.2.1 Combinatorial Machine

A (deterministic) combinatorial machine is an abstract system M=(**X**,**Y**,f), where
- **X** and **Y** are finite sets of recognizable objects (symbols) called the *input set* and the *output set* of M, respectively. These sets are also referred to, respectively, as the *input alphabet* and the *output alphabet* of M.
- f:**X**→**Y** is a function from **X** into **Y** called the *output function* of M.

The work of machine M is described in discrete time, $\nu$, by expression $y_\nu = f(x_\nu)$, where $x_\nu \in$ **X** and $y_\nu \in$ **X** are the input and the output symbols of M, respectively, at the moment $\nu$.

**Example:** **X**={a,b,c}; **Y**={0,1}; f={(a,0),(b,1),(c,0)}. Input 'a' produces output '0', input 'b' produces output '1', and input 'c' produces output '0'.
The pairs of symbols describing function *f* are called *commands*, *instructions*, or *productions*, of machine M.

### 1.2.2 Equivalent Machines

Intuitively, two machines M1 and M2 are *equivalent* if they cannot be distinguished by observing their input and output signals. In the case of combinatorial machines, machines M1 and M2 are equivalent if they have the same input and output sets and the same output functions. Instead of saying that M1 and M2 are equivalent one can also say that machine M1 *simulates* machine M2 and vice versa.

### 1.2.3 Machine Universal with respect to the Class of Combinatorial Machines

A *machine universal with respect to the class of combinatorial machines* is a system MU=(**X**,**Y**,**G**,F), where
- **X** and **Y** are the same as in Section 1.2.1
- **G** is a set of objects called the programs of machine MU.
- F:**X** x **G** → **Y** is a function called the output function of MU. This function is also called the *interpretation* or *decision making* procedure of MU. We will say that this procedure interprets (executes) the program of machine MU or that it makes decisions based on the knowledge contained in this program.

Let MU(g) denote machine MU with a program g∈G. The pair (G,F) satisfies the following *condition of universality*: for any combinatorial machine M=(**X**,**Y**,f) there exists g∈G such that MU(g) is equivalent to M.

**Example:** **X**={a,b}; **Y**={0,1}; and **G** is the set of all possible functions from **X** into **Y**. There exist four such functions: f0={(a,0),(b,0)}; f1={(a,1),(b,0)}; f2={(a,0),(b,1)}; f3={(a,1),(b,1)}, that is **G**={f0,f1,f2,f3}.
In the general case, there exist $n^m$ functions from **X** into **Y**, where n=|**Y**| is the number of elements in **Y**, and m=|**X**| is the number of elements in **X**.

### 1.2.4 Programmable Machine Universal with respect to the Class of Combinatorial Machines

A machine MU from section 1.2.3 is a *programmable* machine universal with respect to the class of combinatorial machines, if **G** is a set of (memory) states of MU, and there exists a memory modification procedure (called programming) that allows one to put machine MU in a state corresponding to any combinatorial machine (from the above class).

### 1.2.5 Learning Machine Universal with respect to the Class of Combinatorial Machines

A programmable machine MU from section 1.2.4 is called a *learning* machine universal with respect to the class of combinatorial machines, if it has a programming (data storage) procedure that satisfies our intuitive notion of learning.
**Note.** In this study learning is treated as a "physical" (biological) rather than a "mathematical" problem. Therefore, we shall not attempt to formally define the general concept of a learning system. Instead, we shall try to design examples of programmable systems that can be programmed in a way intuitively similar to the process of human associative learning.

### 1.2.6 Example: Programmable Logic Array (PLA)

Programmable Logic Array (PLA) is an example of a programmable machine universal with respect to the class of combinatorial machines. The general architecture of a PLA is shown in Figure 1.2. The system has programmable AND-array and programmable OR-array that store, respectively, the input and output parts of commands (productions) of a combinatorial machine. The binary vectors stored in these arrays are represented by the conductivities of fuses (1 is "connected", 0 is "not connected").

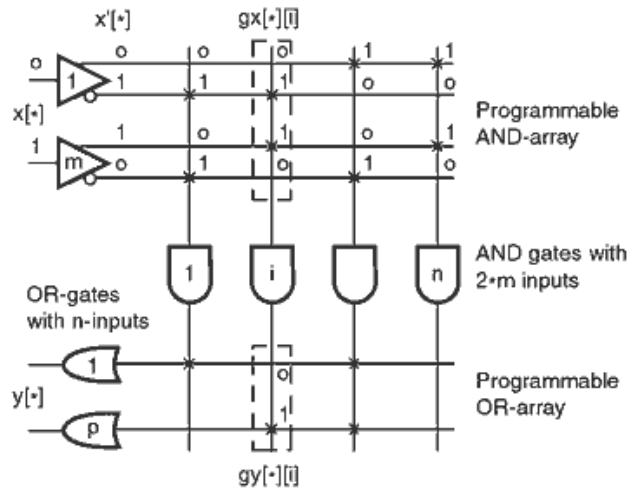

**Figure 1.2 Programmable Logic Array (PLA)**

The input signals, x[*]=(x[1],..x[m]), are m-dimensional binary vectors: x[*]∈**X**={0,1}$^m$ (The '*' substituted for an index denotes the whole set of components corresponding to this index.). Each bit of x[*] is transformed into two bits of vector x'[*] sent to the AND-array: 0→(0,1), and 1→(1,0). This allows one to use 2*m-input AND-gates as match detectors. Unconnected inputs of an AND-gate are equal to 1. Connected inputs are equal to the corresponding components of vector x'[*].
Matrices gx[*][*] and gy[*][*] describe the conductivities of fuses in the AND-array and the OR-array, respectively. It is convenient to think of vectors gx[*][i], and gy[*][i] as data stored in the i-th locations of Input Long-Term Memory (ILTM) and Output LTM (OLTM), respectively.
Using this terminology, the work of a PLA can be described as follows:

1. **Decoding.** Input vector x[*] is compared (in parallel) with all vectors gx[*][i] (i=1,..n) stored in Input LTM.
2. **Choice**. A set of matching locations is found, and one of these locations, call it *i_match* is selected. In the case of a correctly programmed PLA there must be only one matching location.
3. **Encoding.** The vector gy[*][i_match] is read from the selected location of Output LTM and is sent to the output of the PLA as vector y[*]∈**Y**={0,1}$^p$, where *p* is the dimension of output binary vectors (the number of OR-gates).

The concept of PLA was originally introduced in IBM in 1969 as the concept of a Read-Only Associative Memory (ROAM). The term PLA was coined by Texas Instruments in 1970. (See Pellerin, D., et al, 1991.). In Section 1.4, I will show that there is much similarity between the basic topology of PLA and the topology of some popular associative neural networks (Eliashberg, 1993).

### 1.2.7 Finite-State Machine

A (deterministic) finite state machine is a system M=(**X**,**Y**,**S**,$f_y$,$f_s$), where
- **X** and **Y** are finite sets of external symbols of M called (as before) the input and the output sets, respectively.
- **S** is a finite set of internal symbols of M called the state set.
- $f_y$:**X** x **S** → **Y** is a function called the output function of M.
- $f_s$:**X** x **S** → **S** is a function called the next-state function of M.

The work of machine M is described by the following expressions: $s_{v+1}=f_s(x_v,s_v)$, and $y_v=f_y(x_v,s_v)$, where $x_v\in$**X**, $y_v\in$**Y**, and $s_v\in$**S** are the values of input, output, and state variables at the moment *v*, respectively.
**Note.** There are different equivalent formalizations of the concept of a finite-state machine. The formalization described above is known as a *Mealy machine*. Another popular formalization is a *Moore machine*. In a Moore machine the output is described as a function of the next-state. These details are not important for our current purpose.
Practical electronic designers usually use the term *state machine* instead of the term *finite-state machine*.

### 1.2.8 Finite-State Machine as a Combinatorial Machine with a One-Step Delayed Feedback

Any finite-state machine can be implemented as a combinatorial machine with a one-step delayed feedback. The result is obvious from the diagram shown in Figure 1.3.

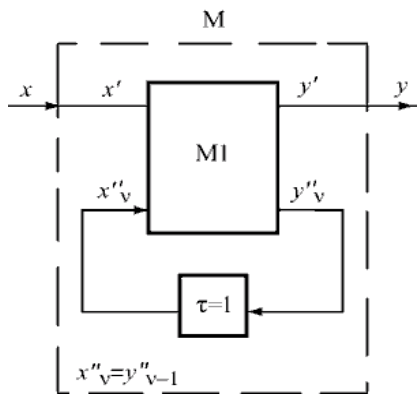

**Figure 1.3 Finite-state machine as a combinatorial machine with a one-step delayed feedback**

In Figure 1.3, M1 is a combinatorial machine and M is a finite-state machine. The one-step delayed feedback $x''_v = y''_{v-1}$ makes $x''_v$ the state variable of machine M. Since one can specify any output function of machine M1 one can implement any desired output and next-state functions for the finite-state machine M.

This result can be naturally extrapolated to programmable (learning) machines universal with respect to the class of finite-state machines. A PLA with a one-step delayed feedback gives an example of a programmable machine universal with respect to the class of finite-state machines.

PLA is often used by logic designers to implement state machines (sequencers). It is also possible to use PROM (Programmable-Read-Only-Memory) and RAM (Random Access Memory) to implement state machines with large numbers of commands, but with relatively small width of input vectors. (The input width is limited by the number of address bits.)

### 1.2.9 Back to Turing's Robot

This section was not intended to serve as a tutorial on finite-state automata and Turing machines. There are many good books (such as Minsky, 1967) which an interested reader should consult for more information. My goal was to illustrate the general concept of an abstract machine and to connect this concept with the notion of a real machine. The main point to keep in mind is that some useful constraints on real machines can be formulated at a rather general system-theoretical level without dealing with specific implementations. When such general constraints exist, it is silly to try to overcome them by designing "smart" implementations. In the same way as it is silly to try to invent a Perpetual Motion machine, in violation of the energy conservation law.

Let us return to the robot shown in Figure 1.1. A Turing machine is a finite-state machine coupled with an infinite tape. Therefore, to be able to simulate any Turing machine, the robot (system (D,B)) must be a learning system universal with respect to the class of finite state-machines. Taking into account what was said in Section 1.2.8 and assuming that the proprioceptive feedback *utter_symbol→symbol_uttered* provides a one-step delay, it is sufficient for system AM to be a learning system universal with respect to the class of combinatorial machines.

Such system is not difficult to design. For example, the PLA shown in Figure 1.2 with the addition of a universal data storage procedure (such as that discussed in Section 1.5) solves the problem. This solution, however, is not good enough for our purpose. We want to implement AM as a neurobiologically plausible neural network model. This additional requirement makes our design problem less trivial and more educational. Solving this problem will put us in a right position for attacking our main problem: the problem of working memory and mental computations (Section 1.6).

## 1.3 Neurocomputing Background

This section provides neurocomputing background needed for understanding the neural model described in Section 1.4.

### 1.3.1 Anatomical Structure of a Typical Neuron

The anatomical structure of a typical neuron is shown in Figure 1.4. The diagram depicts the three main parts of a neuron:

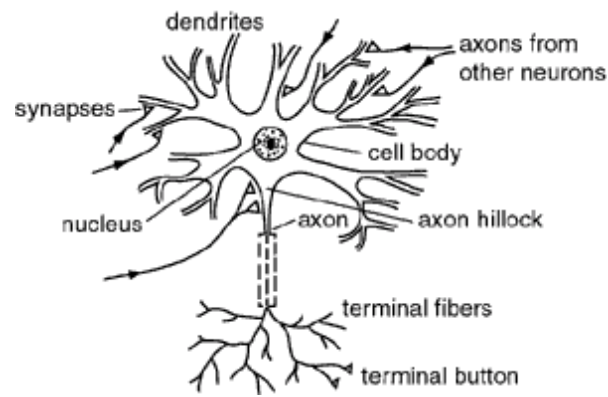

**Figure 1.4 Anatomical structure of a typical neuron**

1. The *cell body* contains the nucleus and all other biochemical machinery needed to sustain the life of cell. The diameter of the cell body is on the order of 10-20µm.

2. The *dendrites* extend the cell body and provide the main physical surface on which the neuron receives signals from other neurons. In different types of neurons, the length of the dendrites can vary from tens of microns to a few millimeters.

3. The *axon* provides the pathway through which the neuron sends signals to other neurons. The signals are encoded as trains of electrical impulses (spikes). Spikes are generated in the area of the axon adjacent to cell body called the *axon hillock*. The duration of a spike is on the order of 2-4msec. The length of some axons can exceed one meter.

A typical axon branches several times. Its final branches, *terminal fibers*, can reach tens of thousands of other neurons. A terminal fiber ends with a thickening called the *terminal button*. The point of contact between the axon of one neuron and the surface of another neuron is called *synapse*.

In most synapses, the axon terminal releases a chemical transmitter that affects protein molecules (receptors) embedded in the postsynaptic membrane. About fifty different neurotransmitters are identified at the present time. A single neuron can secrete several different neurotransmitters. The width of a typical synaptic gap (cleft) is on the order of 200nm. The neurotransmitter crosses this cleft with a small delay on the order of one millisecond.

All synapses are divided into two categories: a) the *excitatory* synapses that increase the postsynaptic potential of the receiving neuron; and b) the *inhibitory* synapses that decrease this potential. The typical resting membrane potential is on the order of -70mV. This potential swings somewhere between +30mV and -80mV during the generation of spike.

Not all axons form synapses. Some serve as "garden sprinklers" that release their neurotransmitters in broader areas. Such non-local chemical messages play important role in various phenomena of activation. See Nicholls, et al, (1992). You may also find useful information in the Web Tutorial at
http://psych.athabascau.ca/html/Psych402/Biotutorials.

It is reasonable to postulate the existence of quite complex computational resources at the level of a single neuron (Kandel, 1968, 1979; Nichols, et al. 1992; Hille, 2001; Eliashberg, 1990a, 2003). In what follows, we won't need this single-cell complexity. A rather simple model of a neuron described below is sufficient for our current purpose. (I must emphasize that a single-cell complexity is needed in more complex models.)

## 1.3.2 Neuron as a Linear Threshold Element

A simple concept of a neuron-like computing element is shown in Figure 1.5. The (a) and (b) parts of this figure illustrate two different graphical representations of this model. The graphical notation (b) will be used in Section 1.4.

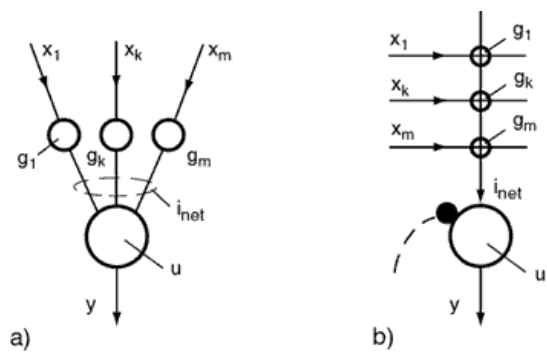

$$i_{net} = \sum_{k=1}^{m} g_k x_k \quad (1) \quad \tau \frac{du}{dt} + u = i_{net} \quad (2)$$

$$\text{if } (u > 0) \ y = u; \ \text{else } y = 0; \quad (3)$$

**Figure 1.5 Simple concept of a neuron-like computing element**

In Figure 1.5a, $x_k$ is the input (presynaptic) signal of the *k-th* synapse, and $g_k$ is the gain (weight) of this synapse. According to Exp (1), the net postsynaptic current, $i_{net}$ is equal to the scalar product of vectors ***g*** and ***x***. In this expression, the excitatory and inhibitory synapses have positive and negative gains, respectively.

In the graphical notation shown in Figure 1.5b, the excitatory and inhibitory synapses are represented by small white and black circles, respectively. To illustrate this agreement, Figure 1.5b shows an inhibitory synapse located on the body of the neuron. (The incoming line and the outgoing line can be thought of as the dendrites and the axon, respectively.)

The dynamics of the postsynaptic potential *u* is described by the first-order differential equation (2). The output signal *y*, described by Exp (3), is a linear threshold function of *u*. For the sake of simplicity the threshold is equal to zero.

### 1.3.3 Using a C-like Language for the Representation of Models

The models we are going to study are too complex for traditional scientific notation. Therefore, I am forced to use some elements of a computer language to represent these models. I want to avoid verbal descriptions unsupported by formalism. Bear with me.

I believe in Herald Morowitz's proposition that "computers are to biology what mathematics is to physics." Trying to avoid computer language and stick with traditional mathematical notation will only prolong one's suffering.

I use a C-like notation assuming that this language is widely known. To be on the safe side, in what follows I explain some of this notation.

- for (i=0;i<n;i++) { *expressions* } means that the expressions enclosed in the braces are computed for i=0,1,...n-1. The post-increment operator "++" increments the value of *i* after each cycle of computations.
- if( b ) *expression1*; else *expression2*; means that if Boolean expression **b** is true compute *expression1*, else compute *expression2*.
- The Boolean expression A==B is true, if A *is equal* to B. The Boolean expression A != B is true if A is *not equal* to B.
- An element of a one-dimensional array is denoted as a[i]. In scientific notation this corresponds to $a_i$. An element of a multi-dimensional array is denoted as m[i][j]..[k], etc. This corresponds to $m_{ij..k}$.

I also use the following "pseudo-scientific" notation:

- The sum a[1]+a[2]+..a[n] is denoted as SUM(i=1,n)(a[i]);
- Exps (1) and (2) from Figure 1.5 will look like this:

    i_net=SUM(k=1,m)(g[k]*x[k]); // (1)

    tau*du/dt+u=i_net; // (2)

    **Note.** Because of the use of multi-character identifiers, I use the multiplication operator '*' explicitly. I use C++ type comments "//" to give expressions the appearance of a computer program.

## 1.4 Designing a Neural Brain for Turing's Robot

This section presents a model of a three-layer associative neural network (Figure 1.6) that can work as a machine universal with respect to the class of combinatorial machines. The network has a PLA-like architecture, and, in fact, has all the functional possibilities of PLA. The use of "analog" neurons (rather than logic gates) gives the network some "extras." The model displays some effect of generalization by similarity and, because of the mechanism of random choice, can simulate, in principle, any probabilistic combinatorial machine (with rational probabilities). A similar model was described in Eliashberg (1967).

The model integrates the following basic ideas:

1. Neuron as a programmable similarity detector. (Rosenblatt, 1962, Widrow, 1962, and others.)
2. Neuron layer with reciprocal inhibition as the mechanism of the winner-take-all choice (Varju, 1965.)
3. Neuron as a programmable encoder (Steinbuch, 1962, and others)

## 1.4.1 Topological Structure of Model

Consider the neural network schematically shown in Figure 1.6. The big circles represent neurons. The small white and black circles denote excitatory an inhibitory synapses, respectively. The incoming and outgoing lines of a neuron represent its dendrites and its axon, respectively.

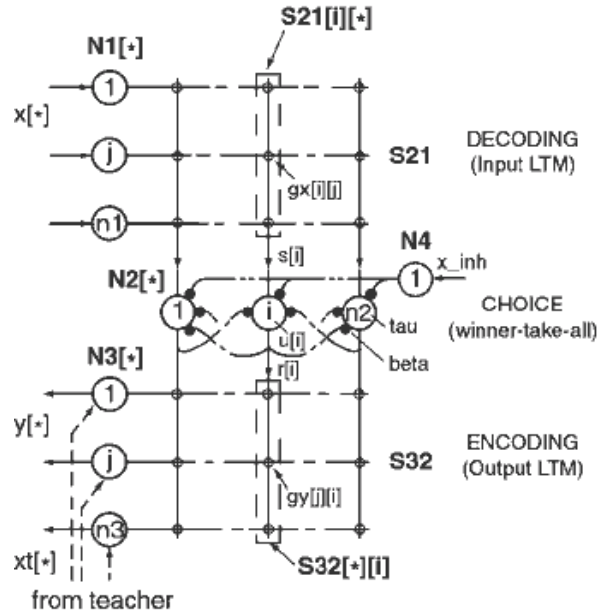

**Figure 1.6 Topological structure of neural network**

The network has four sets of neurons:

- Input neurons N1=(N1[1],..N1[n1])
- Intermediate neurons N2=(N2[1],...N2[n2])
- Output neurons N3=(N3[1],...N3[n3])
- An auxiliary neuron N4

The network has four sets of synapses:

- Input excitatory synapses S21=(S21[1][1],..S21[n2][n1]), where S21[i][j] is the synapse between the axon of N1[j] and a dendrite of N2[i].
- Output excitatory synapses S32=(S32[1][1],..S32[n3][n2]), where S32[j][i] is a synapse between the axon of N2[i] and a dendrite of N3[j].
- Inhibitory synapses S22 in the layer of intermediate neurons N2, that is, synapses from neurons N2 to other neurons N2 (the name S22 is not shown). Every neuron inhibits every other neuron, except itself. With some parameters, such a competition of neurons N2 produces the "winner-take-all" effect. (Figure 1.7 a,b for two different architectures of a winner-take-all layer.)
- Inhibitory synapses S24 between neuron N4 and all neurons from N2. These connections provide a global inhibitory input to layer N2.

**Notation:** I use C-like notation to represent arrays. The '*' substituted for an index denotes the whole set of elements corresponding to this index. In the above description, I should have written N1[*]=(N1[1],...N1[n1]) instead of N1=(N1[1],...N1[n1]), etc. (For the sake of simplicity, I omit "[*]" when such an omission doesn't cause confusion.) In this notation, S21[i][*] is the set of input excitatory synapses of neuron N2[i], and S32[*][i] is the set of output excitatory synapses of this neuron.

**Terminology:**

- The synapse S21[i][j] at the intersection of the axon of neuron N1[j] and the dendrite of neuron N2[i] is referred to as the synapse *from* N1[j] *to* N2[i].

- The graphical notation in which a connection (synapse) is represented by the intersection of two lines will be referred to as "engineering" notation. This type of notation is used in programmable logic devices (PLD). A notation in which a connection is represented by a line between two nodes will be referred to as "connectionist" notation. This notation, borrowed from the graph theory, is commonly used in "connectionist" models. In Section 1.4.6, I will demonstrate advantages of "engineering" notation vs. "connectionist" notation.

### 1.4.2 Functional Model

This section presents a functional model corresponding to the topological model of Figure 1.6. The same topological model may have many different functional models associated with it, so this is just one of such models.

**Notation:**

- *x[\*]=(x[1],...x[n1])* is the vector of output signals of neurons N1: the input vector of the model.
- *y[\*]=(y[1],...y[n3])* is the vector of output signals of neurons N3: the output vector of the model.
- *gx[j][i]* is the gain of synapse S21[i][j]. Vector *gx[\*][i]* will be treated as the contents of the i-th location of the Input LTM (ILTM) of the model. (Note that indices in *gx[j][i]* are transposed as compared to S21[i][j].)
- *gy[j][i]* is the gain of synapse S32[j][i]. Vector *gy[\*][i]* will be treated as the contents of the i-th location of the Output LTM (OLTM) of the model.
- *x_inh* is the output signal of N4. This "nonspecific" signal provides a global inhibitory input to the model.
- *beta* is the absolute value of the gain of a synapse S22[k][i], where *k* is not equal to *i*. The synapse is inhibitory so its gain is equal to -*beta*. The diagonal gains (*k=i*) are equal to zero.
- *s[i]* is the net input current of neuron N2[i] from all neurons N1[\*].
- *r[i]* is the output of neuron N2[i].
- *u[i]* is the postsynaptic potential of neuron N2[i].
- *tau* is the time constant of a neuron N2[i] (the same for all neurons).
- *noise[i]* are the fluctuations of the postsynaptic current of N2[i].
- *t* is continuous time.

**Assumptions:**
- The net input, *s[i]*, of neuron N2[i] from neurons N1[\*] is equal to the scalar product of vectors *x[\*]* and *gx[\*][i]*. This input represents a measure of similarity of input vector *x[\*]* and the vector *gx[\*][i]* stored in the i-th location of ILTM.
- The output, *y[j]*, of neuron N3[j] is equal to the scalar product of gy[j][\*] and r[\*]
- The output, *r[i]* of neuron N2[i] is a linear threshold function of *u[i]* with saturation at *u0*.
- The dependence of postsynaptic potential, *u[i]*, on the net postsynaptic current of N2[i] is described by the first-order linear differential equation with the gain equal to unity and the time constant *tau*.

As mentioned in Section 1.3.3 I use a combination C-like notation and scientific-like notation without subscripts and superscripts, and with variable names (identifiers) containing more than one character:

- The sum of elements *a[i]* from *i=i1* to *i=i2* is denoted as *SUM(i=i1,i2)( a[i] )*.
- The multiplication operator, '\*', is used explicitly.
- C-like control statements are allowed.

The model presented below is described in a C-function-like format, the braces "{ }" indicating the boundaries of the model. I use C++ style comments "//" to give the model an appearance of a computer program. (It is almost a computer program.)

**Model ANN0()** //The abbreviation "ANN0" stands for "Associative Neural Network #0"
{ //Beginning of model ANN0

//**DECODING** (similarity calculation)
for(i=1; i<=n2; i++) s[i] = SUM(j=1,n1)( gx[j][i] * x[j] ); // (1)

//**CHOICE** (competition of neurons N2 via reciprocal inhibition. Expressions (2) and (3))
for(i=1; i<=n2; i++)
{
u[i]=(1-dt/tau)*u[i]+ s[i] - x_inh - beta*SUM(k=1,n2)( r[k] ) + beta*r[i] + noise[i]; // (2) this expression is
//the same as differential equation: tau*du[i]/dt+u[i] = s[i] - x_inh - beta*SUM(k=1,n2)( r[k] ) + beta*r[i] + noise[i];

if( u[i]>0 ) r[i]=u[i];
else r[i]=0; // (3)
}

//**ENCODING** (data retrieval.)
for(j=1; j<=n3; j++) y[j] = SUM(i=1,n2)( gy[j][i] * r[i] ); // (4)

} //End of Model ANN0

### 1.4.3 Two Implementations of the Winner-Take-All Layer

Figures 1.7a and 1.7b show two possible implementations of the winner-take-all layer described by Expressions (2) and (3) from section 1.4.2. The topological model of Figure 1.7a can be referred to as *inhibit-everyone-but-itself* implementation. The topological model of Figure 1.7b can be called *inhibit-everyone-and-excite-itself* implementation. If *alpha=beta*, the two functional models corresponding to these two topological models are mathematically equivalent. If the absolute value of the positive feedback gain, *alpha*, is not equal to the absolute value of the negative feedback gain, *beta*, the model of Figure 1.7b has slightly richer properties than model of Figure 1.7a.

The model of Figure 1.7a was studied in Eliashberg (1967). The model of Figure 1.7b was studied in Eliashberg (1979). In both cases, the systems of differential equations describing the dynamics of these models have explicit solutions for any number of neurons N2 (parameter n2).

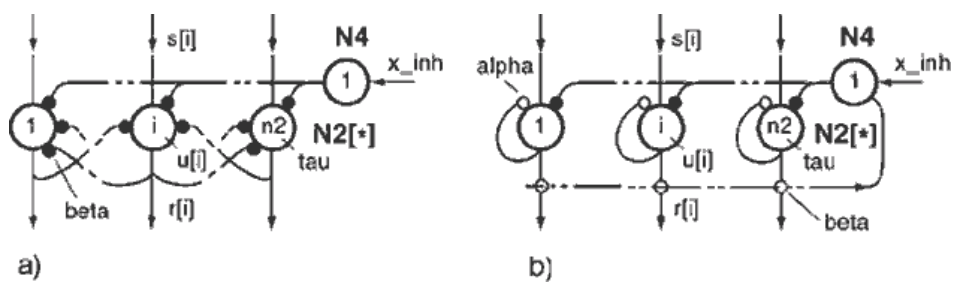

**Figure 1.7 Two implementations of the winner-take-all (WTA) layer**

### 1.4.4 Some Properties of Model ANN0

In what follows I describe some properties of Model ANN0 that can be rigorously proved. (In this study I do not present the proof. The proof can be found in Eliashberg (1979). Model ANN0 doesn't include the description of a learning procedure, so I assume that any desired matrices gx[*][*] and gy[*][*] can be preprogrammed. The pair (gx[*][*],gy[*][*]) will be called the *program* of Model ANN0.

1. Let x[j],y[j],gx[j][i],gy[j][i] ∈ {0,1}. (Inputs, outputs, and gains are binary vectors.) Let *beta > 1, n1=2*m*, and let *n2* be as big as needed.

   For any logic function with *m* inputs and *n3* outputs, F:$2^m \rightarrow 2^{n3}$, there exists a program (gx,gy) such that Model ANN0 with this program implements this function. That is, Model ANN0 can work as a PLA. Each input bit can be encoded as a two bit vector as is done in PLA. For example, (0,1) can represent 0, and (1,0) can represent 1.

2. Let **V** be a set of real normalized positive n1-vectors and let **X** be a finite subset of **V**. Let x[*]∈ **X**. Let **Y** be a finite set of real positive n3-vectors.

   For any function F:**X**→**Y** there exists a program (gx,gy) such that Model ANN0 with this program implements this function.

3. Let the level of *noise[i]* be greater than zero, and let all *noise[i]* (i=1,..n2) be independent random values. Let *0<noise[i] <d*. Let f(x1,x2)=SUM(j=1,n1)( x1[j] * x2[j] ). Let **X** be a finite set of normalized real positive n1-vectors such that for each pair (x1,x2) from **X**, if x1 is not equal to x2, then f(x1,x2) < f(x1,x1)-d. Let **Y** be a finite set of positive n3-vectors.
   Let M=(**X**,**Y**,P) be a probabilistic combinatorial machine with input alphabet **X**, output alphabet **Y**, and the probability function P:**X**x**Y**→[0,1], where P(a,b) is the (conditional) probability that y[*]=b if x[*]=a, where y[*] and x[*] are the output and the input of M, respectively. Let P assume only rational values, *m/n*, where *m* is a non-negative integer and *n* is a positive integer.

   For any machine M there exists a program (gx,gy) such that Model ANN0 with this program implements this machine.

4. To make analog Model ANN0 work as a discrete-time machine (a system with discrete cycles) we need to apply periodic inhibition *x_inh*. This global inhibitory input resets layer N2 after each cycle of random WTA (winner-take-all) choice and prepares it for the next cycle. An analytical solution of equations (2) and (3) from Section 1.4.2 was presented in Eliashberg (1967,1979, and 1988). This solution allows one to understand how layer N2 works. (An attempt to go deeper into this interesting subject would take us too far from the main goal of this study.)

### 1.4.5 Is Model ANN0 Scalable?

- *Is it possible to implement the basic architecture shown in Figure 1.6 with a very large n2 (say, $n2=10^9$)?*

The answer is "Yes". Some plausible topological models providing this answer were discussed in Eliashberg (1979).

### 1.4.6 "Connectionist" Notation vs. "Engineering" Notation

The goal of this section is to show that some of the well known neural network models have essentially the same PLA-like topology as the network of Figure 1.6. They do not look similar to this network because of the use of "connectionist" notation. Switching to "engineering" (PLA-like) notation reveals the similarity.

**Counterpropagation Network (CPN)**

Figure 1.8a shows the topological structure of the CPN as it was presented in Hecht-Nielsen (1987). Figure 1.8b displays another representation of this network borrowed from Freeman, et al. (1991).

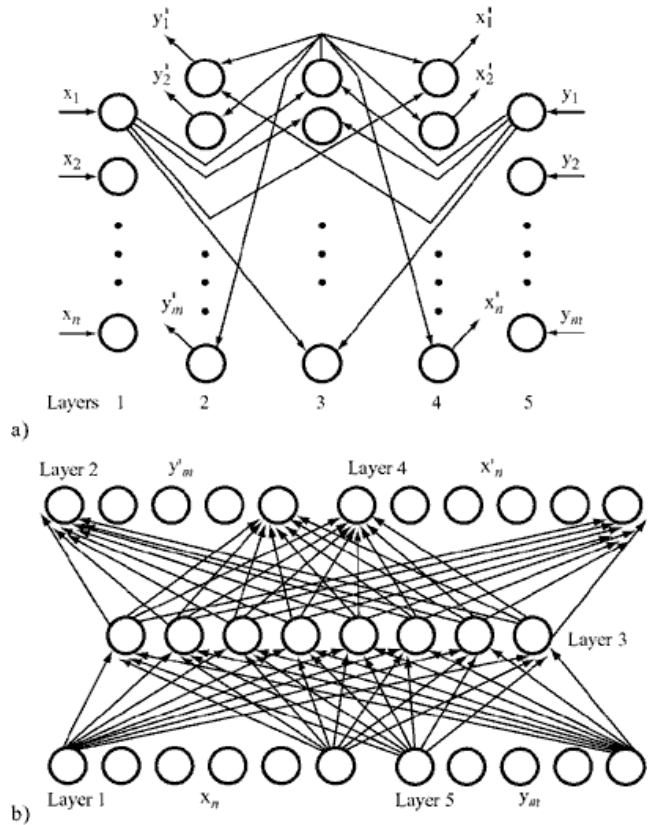

**Figure 1.8 Two connectionist representations of the topological structure of CPN**

Translating these diagrams into the PLA-like "engineering" notation make them look as shown in Figure 1.9. This architecture is remarkably similar to the architecture of the associative neural network of Figure 1.6.

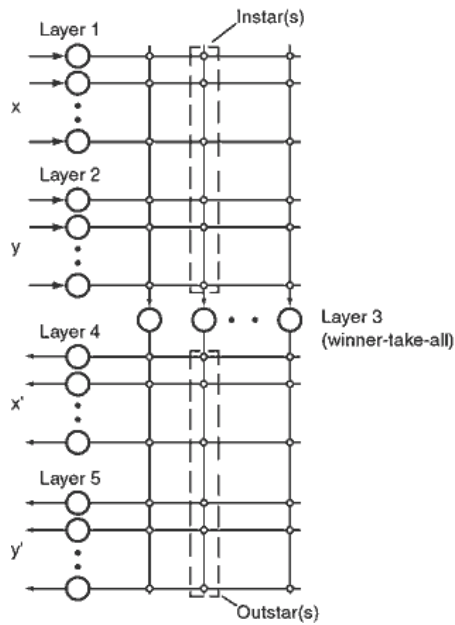

**Figure 1.9 The topological structure of CPN in engineering (PLA-like) notation**

**Adaptive Resonance Theory (ART1) Network**
Figures 1.10 and 1.11 demonstrate a similar transformation in the case of the ART1 network. (Grossberg, 1982)

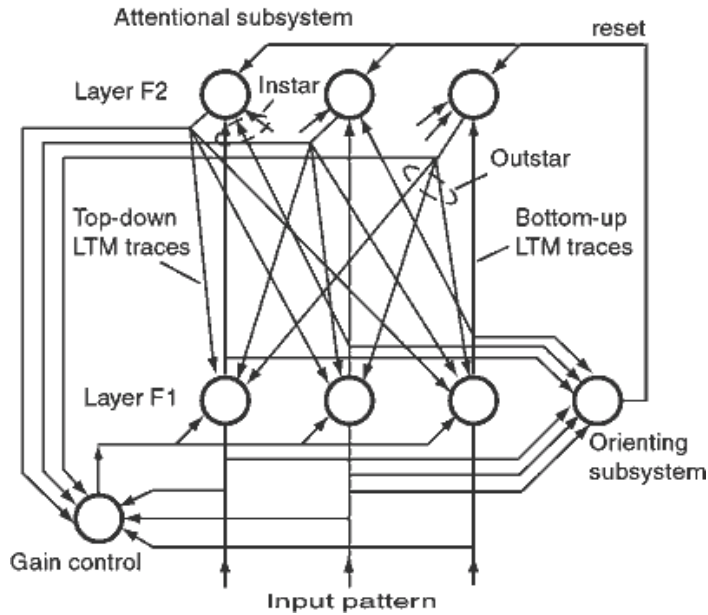

**Figure 1.10 Connectionist representation of ART1 model**

In Figure 1.11, the "bottom-up LTM traces" are similar to the input synaptic matrix S21[*][*] of Figure 1.6 or the AND-array of Figure 1.2.

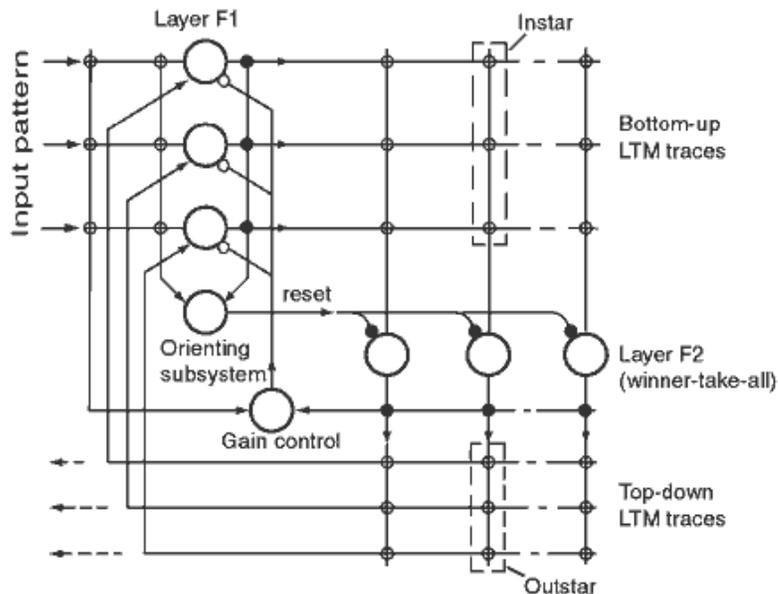

**Figure 1.11 PLA-like representation of ART1 model**

The "top-down LTM traces" are similar to the output synaptic matrix S32[*][*] of Figure 1.6 or the OR-array of Figure 1.2. The other blocks shown in Figure 1.10 are of no interest for our current discussion. Our main issue is how the LTM can be represented in the brain and how the information stored in this LTM can be accessed and retrieved.

## 1.4.7 "Local" vs. "Distributed"

So far we were dealing with a "local" representation of data in the LTM of the network of Figure 1.6: "one neuron - one memory location." In Feldman's terminology this "local" approach is referred to as the "grandmother cell" approach. (This, of course, is just an easy-to-remember metaphor. There is no single cell in the brain representing one's grandmother.)

- *What happens if we reduce the strength of reciprocal inhibition in Model ANN0?*

With *beta<1*, layer **N2** no longer works as a winner-take-all mechanism. Instead, it produces effect of contrasting and selects a set of several (more than one) locations of Output LTM (call it **ACTIVE_SET**). A superposition of vectors gy[*][i] (with i∈ **ACTIVE_SET**) is sent to the output *y[*]* of Model ANN0. We can no longer treat this model as "local" associative memory.

Let us assume that the output of a neuron is a sigmoid function (Figure 1.12b) of its postsynaptic potential (instead of the linear threshold function used in Model ANN0). Let us also completely turn off reciprocal inhibition by setting *beta=0*. Model ANN0 becomes a traditional three-layer "connectionist" neural network shown in "connectionist" notation in Figure 1.12a.

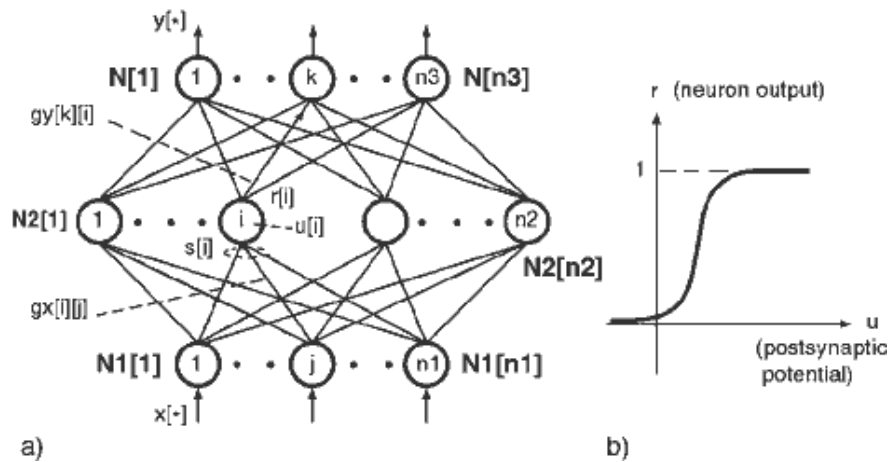

**Figure 1.12 Model ANN0 reduced to a three-layer "connectionist" network by turning off reciprocal inhibition (*beta=0*)**

In the current paper we are interested only in the "local" case corresponding to *beta>1*. A "distributed" case (*beta<1*) becomes important in the models with hierarchical structure of associative memory (Eliashberg, 1979).

## 1.5 Associative Neural Networks as Programmable Look-up Tables

This section discusses a discrete-time counterpart of the continuous-time neural model described in the previous section. It is convenient to view this discrete-time system as a programmable look-up table (LUT). This LUT is transformed into an LUT with "dynamical bias" by introducing the states of "residual excitation" (E-states). This leads to the concept of a primitive E-machine. (Eliashberg, 1967, 1979, 1981, 1989.)

### 1.5.1 Model AF0

Let as assume that the input vectors of Model ANN0 are changing step-wise with a time-step *ΔT>>tau*. Let *beta>1*, so the layer **N2** performs a random winner-take-all choice. Let *x_inh* provide a periodic inhibition needed to reset the layer after each step. The exact values of parameters are not important for the current discussion.

In the above step-wise mode of operation, the network of Figure 1.6 can be replaced by the programmable "look-up table" schematically shown in Figure 1.13. The functional model presented below is referred to as Model AF0 (Associative Field #0.) The model is described as a composition of the following blocks:

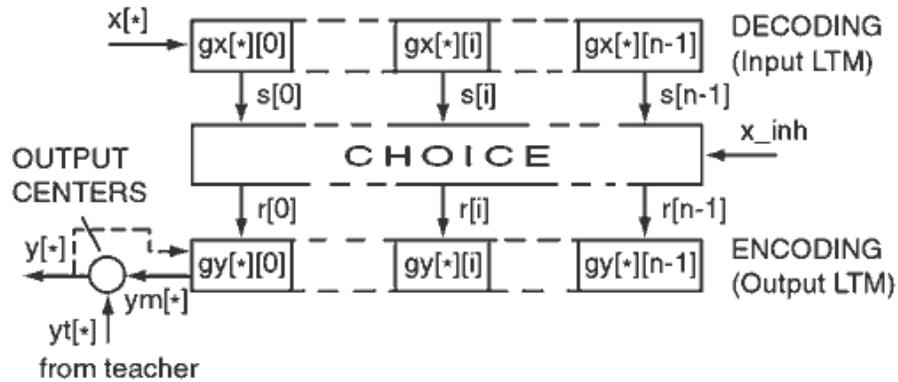

**Figure 1.13 Associative neural network as a look-up table (Model AF0)**

- **DECODING**. This block compares the input vector x[*] with all vectors gx[*][i], i=0,..n-1, stored in Input LTM. As a result of this parallel comparison the front of similarity, s[i] (i=0, n-1) is calculated. In Model ANN0 (Figure 1.6) this block is implemented by synaptic matrix **S21** and by the summation of the postsynaptic currents in neurons **N2**.

- **CHOICE**. This block transforms its input front s[*] into its output front r[*]. In the simplest case it performs a random equally probable choice of a single component *i_read* corresponding to the position of one of the maxima of front s[*]. In Model ANN0 this block is implemented by the layer **N2**.

- **ENCODING**. This block produces the output vector ym[*] as a result of interaction of front r[*] with the data, gy[*][*], stored in the Output LTM. In the simplest case ym[*]=gy[*][i_read]. That is the output vector is read from the location of Output LTM selected by the block **CHOICE**. In Model ANN0 this block is implemented by synaptic matrix **S32** and neurons **N3**.

- **OUTPUT CENTERS**. This block works as a multiplexer: if(select==0) y[*]=yt[*]; else y[*]=ym[*];

- **LEARNING**. This block is not shown in Figure 1.13. It calculates the next values of gx[*][*] and gy[*][*]. In Model ANN0 this block was not described at all. In this model it is described in procedural (algorithmic) terms without any neural interpretation. Possible neural implementations of different learning algorithms will be discussed in Chapter 2.

**Notation:**
As before, I use a C-like notation mixed with scientific-like notation. I use special notation for two important operations: *select a set*, and *randomly select an element from a set*.

1. **A**:={a : P(a)} select the set of elements *a* with the property P(a). I use Pascal-like notation ":=" to emphasize the dynamic character of this operation.

2. a:∈ **A** select an element *a* from the set **A** at random with equal probability.

**Note.** I want to remind the reader that all models in this study are aimed at humans. For the purpose of computer simulation, it is easy to replace operations ":=" and ":   " with valid C or C++ functions.

**Model AF0()**
{ //Beginning of Model AF0

//**DECODING**

for(i=0;i<n;i++) s[i]=Similarity(x[*],gx[*][i]); // (1) compare input vector with all vectors in Input LTM

//**CHOICE** Exprs (2) and (3)

**MAXSET**:={i : s[i]=max(s[*]) }; // (2) select the set of locations with the maximum value of s[i]

i_read :∈ **MAXSET**; // (3) randomly select a winner (i_read) from **MAXSET**

if(s[i_read] > x_inh) ym[*] = gy[*][i_read]; else ym[*]=NULL; // (4) read output vector

//from the selected location of Output LTM. "NULL" stands for "no signals".

//**OUTPUT CENTERS**

if(select==0) y[*]=yt[*]; else y[*]=ym[*]; // (5) if( select==0) the output is from teacher, else it
//is read from memory

//**LEARNING**

if (learning_enabled) {gx[*][i_write]=x[*]; gy[*][i_write]=y[*]; i_write++;} // (6)

//if learning is enabled record X-sequence and Y-sequence in Input LTM and Output LTM, respectively.

}// End of Model_AF0

**Note.** Don't be discouraged by the simplicity of the "dumb" learning algorithm described by Exp. (6). Theoretically, it is the most universal and powerful learning procedure possible (it stores all available input and output experience just in case). Practically, it is not too bad because the size of the required memory grows only linearly with time. Since the memory is addressed by content the decision making time doesn't increase much with the increase of the length of the recorded XY-sequence. (Keep in mind that the presently popular "smart" learning algorithms throw away a lot of information available for learning.) It is easy to improve this "dumb" learning algorithm to make it less "memory hungry." The first obvious improvement is "selection by novelty". In the program EROBOT the user can select one of two learning modes: 1) storing all XY-sequence, 2) storing new XY- pairs.

### 1.5.2 Correct Decoding Condition

We didn't specify similarity function. Any combination of input encoding (set **X**) and similarity function (Similarity() ) will work as long as this combination satisfies the following *correct decoding condition*.

**DEFINITION.**
Let **X** be the set of allowed values of input variable x[*], and let f:**X** x **X** → **R** be a function from **X** x **X** into the set of real numbers (usually the set of non-negative numbers with some upper limit). We will say that set **X** satisfies *correct decoding condition* with the similarity function *f*, if

$\forall$ a,b ∈ **X** if(a!=b) then f(a,a) > f(a,b)        (1)

where

- "$\forall$ a,b ∈ **X**" means "for all a ∈ **X** and for all b ∈ **X** "
- "a!=b" means " a *is not equal* to b "

Informally, the *correct decoding condition* (1) means that any allowed input vector must be "more similar" (closer) to itself than to any other allowed input vector.
**EXAMPLE**
- The set of normalized real vectors satisfies correct decoding condition with the similarity function in the form of the scalar product.

### 1.5.3 What Can Model AF0 Do?

The work of the "psychological" Model AF0 is much easier to understand than the work of "neurobiological" Model ANN0. Nevertheless, the information processing (psychological) possibilities of Model AF0 are essentially the same as those of Model ANN0 (with *beta>1* and with the input signals changing step-wise with the time step much larger than *tau*).

We no longer need to talk about neurons and synapses, and can treat Model AF0 as a programmable look-up table with some effect of "generalization by similarity". It is heuristically important, however, to keep in mind the relationship between Model AF0 and Model ANN0.

In what follows I describe some properties of Model AF0 without a proof. (The proof was given in Eliashberg, 1979.) I assume that the input set **X** and the Similarity() function are selected in such a way that the *correct decoding condition* from Section 1.5.2 is satisfied.

1. The process of training during which the teacher can produce any desired XY-sequence will be referred to as *XY-training*. Experiments of XY-training are often called experiments of *supervised learning*.

2. As in Model ANN0, the pair (gx[*][*],gy[*][*]) is called the *program* of Model AF0. (When it doesn't cause confusion, I use notation *gx* and *gy* instead of *gx[*][*]* and *gy[*][*]*, respectively.) It is easy to see that any program (gx,gy) can be created in the LTM of Model AF0 via XY-training. (If *learning_enable* is TRUE, the XY-sequence is recorded as the program.)

3. Model AF0 can be trained to simulate any probabilistic combinatorial machine with rational probabilities. That is, AF0 is a learning system universal with respect to the class of (probabilistic) combinatorial machines.

4. Let x[*]=(x1[*],x2[*]), y[*]=(y1[*],y2[*]). Let *v* denote discrete time (step number). Let us introduce one step delayed feedback x2[*]$_{v+1}$=y1[*]$_v$ as shown in Figure 1.14.

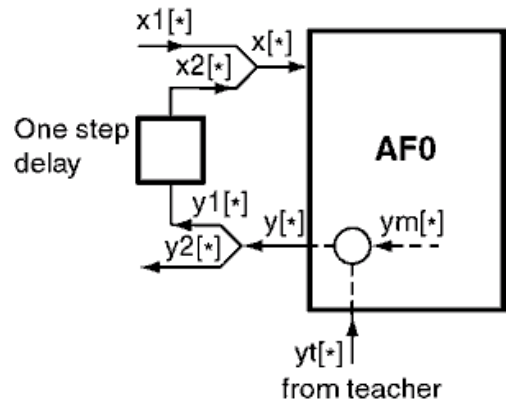

**Figure 1.14 Transforming Model AF0 into a learning system universal with respect to the class of finite-state machines**

Let us use x1[*] as input variable, y2[*] as output variable, and x2[*] as state variable. It is easy to prove that the resulting learning system is universal with respect to the class of probabilistic finite-state machines (with rational probabilities).

### 1.5.4 "Neurobiological" vs. "Psychological" Models

It is useful to compare the general structure of the "neurobiological" model ANN0 from Section 1.4.2 with that of the "psychological" model AF0 from Section 1.5.1.

**Terminology and notation**

- A *neurobiological time step* dt is a time step sufficiently small to correctly simulate neurobiological phenomena. The exact value of dt is of no importance for the current discussion. (One can suggest, for example, that dt<1usec would be sufficiently small.)
- A *psychological time step* $\Delta t$ is a time step sufficiently small to correctly simulate psychological phenomena. Psychological time step is much greater than neurobiological time step, that is $\Delta t \gg dt$. (It is reasonable to assume that $\Delta t$ can be as big as 10msec or even bigger.)

**General structure of model ANN0**
The work of neurobiological model ANN0 can be described in the following general form:

$$y_t = f_y(x_t, u_t, g_t) \qquad (1)$$
$$u_{t+dt} = f_u(x_t, u_t, g_t) \qquad (2)$$

where
- $x_t$, and $y_t$ are, respectively, the value of input and output vector of Model ANN0 at time t.
- $u_t$ is the value of the array of postsynaptic potentials of neurons N2 at time t. The only state of STM of model ANN0.
- $g_t$ is the state of (input and output) LTM of Model ANN0.
- $f_y$, and $f_u$ are, respectively, the output function and the next-state function

**Note**. Variables x_inh and noise are omitted for simplicity.

**General structure of model AF0**
The work of psychological model AF0 can be described in the following general form:

$$y_t = F_y(x_t, g_t) \qquad (3)$$
$$g_{t+\Delta t} = F_g(x_t, y_t, g_t) \qquad (4)$$

where
- $x_t, y_t$ and $g_t$ have the same meaning as in model ANN0
- $F_y$ is the output function of Model AF0
- $F_g$ is the next-LTM-state function of model AF0 (also called learning or data storage procedure of this model).

**Comparison**
1. The output function $F_y$ of model AF0 is much simpler than the output function $f_y$ of model ANN0. $F_y$ is a result of many steps of work of model ANN0.
2. The state of "neurobiological" STM of model ANN0 (state *u*) is not needed in psychological model AF0.
3. It was easy to introduce learning algorithm in model AF0. (It would be more difficult to do so in model ANN0).

## 1.5.5 Expanding the Structure of Model AF0 by Introducing E-states

Because of its simplicity, model AF0 has room for development. The most important of such developments is the introduction of "psychological" STM. The term "psychological" means that the duration of this memory must be longer than the psychological time step $\Delta t$. The states of such memory are referred to in this study as the states of "residual excitation" or E-states. Let us add E-states to model AF0.

$$y_t = F_y(x_t, e_t, g_t) \qquad (5)$$
$$e_{t+\Delta t} = F_e(x_t, e_t, g_t) \qquad (6)$$
$$g_{t+\Delta t} = F_g(x_t, y_t, e_t, g_t) \qquad (7)$$

- *What can be a neurobiological interpretation of E-states?*

The possibility of connecting the dynamics of the postulated phenomenological E-states with the statistical dynamics of the conformations of protein molecules in neural membranes is discussed in Eliashberg, (1989, 1990a, 2003).

- *What can be achieved by the introduction of E-states?*

Here are some possibilities associated with E-states (Eliashberg, 1979).

1. Effect of read/write working memory without sacrificing the ability to store (in principle) the complete XY-experience. An example of a primitive E-machine described in the next section illustrates this effect. This model is used as system AS in the universal learning robot shown in Figure 1.16 (Sections 1.6 and 1.7).

2. Effect of context-dependent dynamic reconfiguration. The same primitive E-machine can be transformed into a combinatorial number of different machines by changing its E-states. No reprogramming is needed!

3. Recognition of sequences and effect of temporal associations.

4. Effect of "waiting" associations and simulation of stack (with limited depth). This leads to the possibility of calling (and returning from) "subroutines."

5. Effect of imitation. A sensory image of a sequence of reactions "pre-activates" (pre-tunes) this sequence. This effect allows the synthesis of complex motor reactions by presenting their sensory images. One can start with "bubbling" and create complex sequences. This explains how complex reactions can be learned without the teacher's acting directly on the learner's motor centers (as it is done in the simple robot discussed in this chapter).

### 1.5.6 Model AF1: An Example of Primitive E-machine

The general structure of Model AF1 is shown in Figure 1.15. The model includes the following blocks:

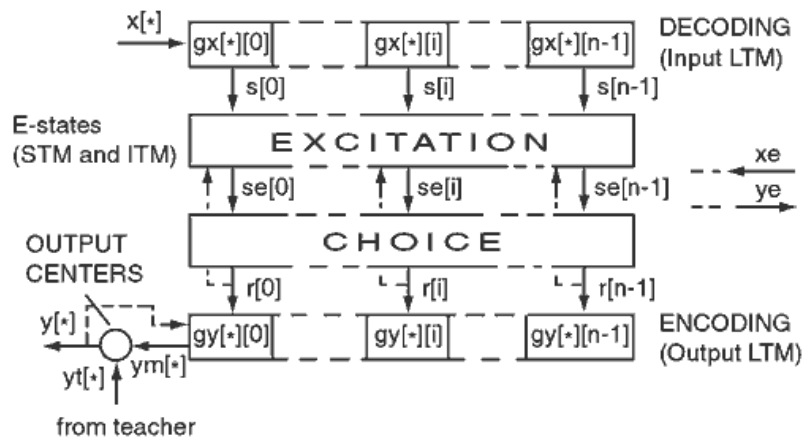

**Figure 1.15 The simplest architecture of a primitive E-machine**

- **DECODING**. This block is similar to the corresponding block of Model AF0.

- **EXCITATION**. This is a new block. It receives similarity front, s[*], as its input, and produces the front of "biased similarity", se[*], as its output. The "bias" is associated with the E-states mentioned in the previous section. In the functional model described below the work of this block is described by two procedures:

    1. **BIAS** that calculates similarity, se[*], biased by the effect of "residual excitation." Coefficients *ba* and *bm* describe *additive* and *multiplicative* biasing effect, respectively.

2. **NEXT E-STATE PROCEDURE** that calculates the next E-state. In Model AF1 there is only one type of E-states, *e[*]*. All components e[i] (i=0,n-1) have the same time constant of "discharge", *tau*. In this model, the "charging" of *e[i]* is instant, so no time constant is specified.

   **Note.** In more complex models of primitive E-machines one can have many different types of E-states with different types of dynamics. This simple model is sufficient for our current purpose. As will be explained in the next section, in spite of its simplicity, Model AF1 produces an effect of read/write "symbolic" working memory that allows the robot of Section 1.6 to learn to perform mental computations. Once the main idea is understood, this critically important effect can be produced in many different ways.

- **CHOICE** is similar to that of Model AF0.

- **ENCODING** is similar to that of Model AF0.

- **OUTPUT CENTERS** is the same as in Model AF0.

- **NOVELTY DETECTION**

   **Note.** At this point we don't care about a specific implementation of Novelty() function. It is sufficient to know that such "nonspecific" computational procedures can be naturally integrated into models of primitive E-machines. Methodologically, it is a separate problem how to implement such computational procedures in neural models.

- **LEARNING** is the same as in Model AF0 with the addition of selection by novelty. There are three modes of learning:

   1. Recording all XY-experience. This mode is activated when *learning_enable* is equal to zero.

   2. Recording of novel X→Y associations. This mode is activated when *learning_enable* is equal to one.

   3. No learning. This mode is in effect if *learning_enable* is different from zero and one.

**Model_AF1()**
{ //Model AF1 begins

//**DECODING**
for(i=0;i<n;i++) s[i]=Similarity(x[*],gx[*][i]); // (1) compare input vector with all vectors in Input LTM

//**BIAS**
for(i=0;i<n;i++) se[i]=s[i]*(1+bm*e[i])+ba*e[i]; // (2) calculate "biased" similarity se[*]

//**CHOICE** Exprs. (3) and (4)
**MAXSET** := {i : se[i]=max(se[*]) }; // (3) Select the set of locations with the maximum value of se[i]

i_read : ∈ **MAXSET**; // (4) Randomly select a winner (i_read) from **MAXSET**

//**ENCODING**
if(se[i_read]>x_inh) ym[*] = gy[*][i_read]; else ym[*]=NULL; // (5) read output vector from the selected
//location of Output LTM

//**OUTPUT CENTERS**
if(select==0) y[*]=yt[*]; else y[*]=ym[*]; // (6) if( select==0) the output is from teacher, else it
//is read from memory

//**NOVELTY DETECTION**
x_is_new=Novelty(x[*],gx[*][*]); // (7) x_is_new=TRUE, if there is no gx[*][i] "similar enough" to x[*]

//**NEXT E-STATE PROCEDURE**
for(i=0;i<n;i++) e[i]=(1-1/tau)*e[i]; // (8) residual excitation decays with time constant *tau*

if(!x_is_new) e[i_read]=1; // (9) the winner is biased (if the input is not new)

//**LEARNING**
if (learning_enable ==0 || learning_enable==1 && x_is_new)
{gx[*][i_write]=x[*]; gy[*][i_write]=y[*]; // (10) XY-association is recorded

e[i_write]=1; // (11) the "recording neuron" is biased

i_write++;} // (12) write pointer is incremented

}// End of Model_AF1

**Note.** The program EROBOT uses a slightly more complex data storage procedure than that described by Exp. (12). To allow the user to erase and reuse parts of robot's memory, the recording is done in the first "empty" location. Non-empty locations are skipped. Also Exp. (6) has a parameter that allows the user to switch between "teacher" mode and "memory" mode.

## 1.6 Robot That Learns to Perform Mental Computations

This section enhances the structure of the cognitive model shown in Figure 1.1 to give the robot an ability to learn to perform mental computations.

### 1.6.1 General Structure

Compare the cognitive model shown in Figure 1.16 with the model shown in Figure 1.1. The model of Figure 1.16 has the following enhancements:

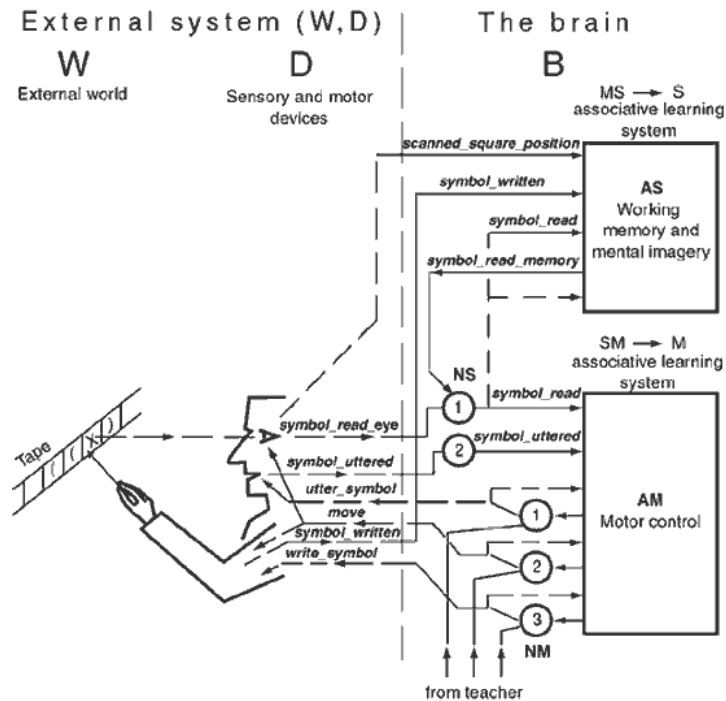

**Figure 1.16 Robot that learns to perform mental computations**

- There is a new associative learning system AS that forms Motor,Sensory→Sensory (MS→S) associations. The goal of this system is to simulate the external system (W,D) as it appears to system AM. The interaction between "processor" AM and "memory" AS creates a universal computing architecture that can perform, in principle, any computations.

  **Note.** The primitive E-machine (Model AF1) described in Section 1.5.6 is used as system AS. The trivial primitive E-machine (model AF0) from Section 1.5 is used as system AM. (A trivial primitive E-machine is an E-machine without E-states.) The effect of a read/write working memory buffer in system AS is achieved automatically as an implication of E-state, e[*]. In the current model, system AM doesn't need E-states.

- To be able to simulate the external read/write memory device (the tape) system AS needs two additional inputs: *scanned_square_position* that serves as "memory address," and *symbol_written* that works as "data." No counterpart of the "write_enable" control signal is needed.

- Sensory centers NS1 that served no useful purpose in the model of Figure 1.1 now serve as a switch. If the robot's eye is open, the output of NS1 is equal to the output of the eye. Otherwise, the output of NS1 is equal to the output of AS. That is, when the eye is closed the AM automatically gets its input from AS. In the program EROBOT the opening and closing of the eye is controlled by user. In a more complex model, this can be done by system AM.

### 1.6.2 Model WD1: Functional Model of External System

To avoid ambiguity, in what follows I present an explicit description of the work of external system (W,D). This description is referred to as Model WD1.

**Inputs** The inputs of system (W,D) are the motor outputs of centers NM.

- *utter_symbol* $\in$ **Q** causes the robot to utter a symbol representing the internal state of a Turing machine, where **Q** is the set of internal symbols. This set includes symbol "H" that causes the Turing machine to halt.

- *move* $\in$ {L,S,R} causes the robot to move one step to the left, stay in the same square, and move one square to the right, respectively.

- *write_symbol* $\in$ **S** causes the robot to write the corresponding symbol into the scanned square (the old symbol is replaced), where **S** is the set of external symbols of the Turing machine.

**States**

- *tape[i]* $\in$ **S** is the symbol in the i-th square of the tape, where i=0,1,2,....

- *i_scan* $\in$ {0,1,2,...} is the position of the scanned square

- *symbol_uttered* is the one-step-delayed input *utter_symbol*

- *symbol_written* is the one-step-delayed input *write_symbol*

**Outputs**

- *scanned_square_position* is the same as the state *i_scan*

- *symbol_read_eye* is the symbol read from the scanned square.

- *symbol_written*, and *symbol_uttered* are the same as the corresponding states.

**Model WD1()**
{ //Beginning of Model WD1

//**OUTPUT PROCEDURE**
symbol_read_eye=tape[i_scan]; // (1) read symbol from the scanned square

scanned_square_position=i_scan; // (2) scanned square position

//**Note:** outputs *symbol_written* and *symbol_uttered* are described in the next-state procedure

//**NEXT-STATE PROCEDURE**
symbol_uttered=utter_symbol; // (3) symbol uttered at the previous step

symbol_written=write_symbol; // (4) symbol written at the previous step

tape[i_scan]=write_symbol; // (5) symbol written at the previous step

if(move=='L') i_scan--;
if(move=='R') i_scan++; // (6) move to the next square

} // End of Model WD1

### 1.6.3 Describing Coordinated Work of Several Blocks

To get a complete working functional model of the whole cognitive system (W,D,B) shown in Figure 1.16 one needs to connect all blocks as shown in this figure. For simplicity, I do not formally describe these connections assuming that they are sufficiently clear from Figure 1.16.

In this simple case, the descriptions of blocks NM and NS (call them nuclei) were included in the descriptions of blocks AM and AS (Models AF0 and AF1.) In more complex cases it is more convenient to describe nuclei as separate blocks. Note that signals from eye (*symbol_read_eye*) play the same role for NS as signals from teacher play for NM.

At this point, I also do not explicitly describe timing details associated with coordinated work of blocks. It is easy to solve such timing problems in computer simulation by doing computations associated with different blocks in a right order. Such timing details, however, become critically important when one addresses the problem of "analog neural implementation" of complex E-machines composed of several primitive E-machines, and nuclei, and including various feedback loops. This very interesting and complex neurodynamical problem will not be discussed in this paper. There is a vast unexplored world of sophisticated neurodynamical problems.

## 1.7 Experiments with EROBOT

### 1.7.1 How to Get the Program

The best way to understand how the robot of Figure 1.16 works is to experiment with the program EROBOT. The program is available at www.brain0.com/software.html. In the following sections I assume that you have acquired this program, and have it running on your computer.

### 1.7.2 User Interface

When you run the program for the first time, four windows are shown on the screen. The windows are resizable so you can rearrange them to your liking. If you want to preserve this new arrangement go to **File** menu and select **Save as default** item. Next time the program will start with your arrangement. For now leave the windows as they are displayed.
The windows have the following titles:
- **EROBOT.EXE** This is the program window. This window has the menu bar with the following menu titles: **File**, **Examples** and **Help**. **File** menu allows you to save and load projects. **Examples** menu has five examples that demonstrate how the program works and help you learn the user interface. The interface is simple and intuitive, so not much help is needed to master it.

- **WORKING MEMORY AND MENTAL IMAGERY (AS and NS)** This window corresponds to the Associative Filed **AS**, and to the sensory nucleus **NS** of Figure 1.16. This Associative Field (primitive E-machine) forms MS→S associations and learns to simulate the external system (W,D). The long table on the right displays the contents of the Input and Output LTM of **AS**. The shorter table on the left displays the input and output signals. You can edit the names of these signals by clicking on these names. The upper control **from: tape memory** determines where the input signals are coming from. The lower control **learn: all new none** switches the learning mode. Click on the desired mode to activate it. The red color corresponds to the active mode. Symbols can be entered in the right table and the table can be scrolled. Click the left mouse button in a square to place the yellow cursor in this square. You can now enter the desired character from the keyboard. To empty a square press the **space bar**. Try **Backspace** and **Del** keys to see how they work. The table can store up to 1000 associations. To scroll the table toward higher addresses press the **F12** key or move the yellow cursor by pressing the → key. To scroll toward lower addresses press the **F11** key or move the yellow cursor by pressing the ← key. Press the **End** key to go to the end of the table and the **Home** key to return to the beginning of the table.
  **Note.** The keys work only when the window is selected by clicking the left button inside the table.
- **MOTOR CONTROL (AM and NM)** This window corresponds to the Associative Field **AM** and the motor nucleus **NM** of Figure 1.16. This Associative Field forms SM→M associations and learns to simulate the teacher. Editing and scrolling functions, and **from** and **learn** controls are similar to those of **AS** window. When **from** control is in **teacher** position the user can enter symbols in the lower three squares (below the red line) of the right column of the input/output table. Click the left mouse button in one of these squares to position the yellow cursor in this square. You can now enter a symbol from the keyboard. When **from** control is in **memory** position you cannot enter symbols in these squares, because these "motor" symbols are read from memory. You can force the robot's motor reactions only in **teacher** mode.
- **EXTERNAL SYSTEM (W,D): Tape, Eye, Hand, and Speech organ** This window corresponds to the external system **(W,D)** of Figure 1.16. It displays the current state of the tape (the top white row) and the tape history (the gray area below the current tape). The tape can be up to 1000 squares long. The tape history stores 199 previous states of the tape, so you can trace the performance of your Turing machine during the last 200 steps. To edit the tape click left mouse button on the desired square. The yellow cursor is positioned in this square. You can now enter a symbol from the keyboard. The green cursor represents the *scanned square*. To position this cursor click the right mouse button in the desired square. The yellow cursor can be positioned in the history area, but only the current (white) tape can be edited. To scroll the tape left and right press **F12** and **F11** keys, respectively, or move the yellow cursor by pressing the → or ← keys. To scroll the history table up and down press the **PgUp** and **PgDn** keys or move the yellow cursor up and down by pressing the ↑ and ↓ keys. The **Home** key returns the user to the beginning of the current tape. The **End** key displays the end of the tape. The leftmost column displays discrete time (the step number). The next columns display the command (SM→M association) executed by block **AM** at this step.

The buttons in **AS** and **AM** windows perform the following functions:

- The buttons **Clr S,E** in blocks **AS** and **AM** clear the Similarity, *s[*]* array and E-state, *e[*]*, array displayed in the bottom right part of the window. The *s[winner]* is displayed in read, *s[i]* for other locations is displayed in green, and *e[i]* is displayed in magenta. (In this model the bias in block **AM** is turned off, so there is no E-state display.) The control tau = 50 displays the time constant of decay for *e[i]* in block **AS**. This value can be changed by clicking on the number.
- The buttons **Clr G** in blocks **AS** and **AM** clear the right table displaying the associations stored in LTM.
- The button **Clr TH** in **AM** window clears the Tape History in the **(W,D)** window. The button **Clr T** clears the current tape (the upper white row) in **(W,D)**.
- The **Init** and **Step** buttons in **AM** window control the work of robot. Pressing the **Init** button performs a one step of computations without affecting the state of tape and without incrementing time. This button is used to put the robot in the desired initial state. Pressing the **F2** key produces the same effect. Pressing the **Step** button or the **F1** key performs a complete cycle of one-step computations. The state of tape and the time are changed, and the tape history is scrolled.

### 1.7.3 Calculation of Similarity and Bias

To understand the display of *s[i]* and *e[i]* in the lower right part of **AS** and **AM** windows you should know how these values are calculated. The similarity, *s[i]*, is computed as follows:

s[i]=SUM(j=0,nx-1)(truth(x[j]==gx[j][i] && x[j]!=' ')/SUM(j=0,nx-1)(truth(x[j]!=' ')

where
*truth(TRUE)=1* and *truth(FALSE)=0*. The blank (' ') character represents "no signals." If the SUM in the denominator is equal to zero, then *s[i]=0*. It is easy to see that the maximum value of similarity is 1. Many other Similarity() functions would work as well.

Model AF1 described in Section 1.5.6 is used as **AS** and **AM**. In the case of **AS**, the "multiplicative" biasing coefficient *bm=0.5* and the "additive" biasing coefficient *ba=0.0* (see Exp. (2) of Section 1.5.6). In the case of **AM** *bm=ba=0* (no bias). Accordingly, the value of time constant *tau* is needed only in block **AS** and the E-state front, *e[*]* is displayed only in this block (magenta).

### 1.7.4 Example 1: Computing with the Use of External Tape

Go to **Examples** menu and select Example 1. A set of 12 commands representing a Turing machine is loaded in the LTM of block **AM**. This Turing machine is a parentheses checker similar to that described in Minsky (1967). The tape shows a parentheses expression that this Turing machine will check.
Symbols **A** on both sides of the parentheses expression serve as the delimiters indicating the expression boundaries. The green cursor indicating the scanned square is in square 1. Note that *symbol_uttered='0'* showing that the Turing machine has initial "state of mind" represented by symbol '0'.

Push **Step** button or **F1** key to see how this machine works. The machine reaches the Halt state 'H' and writes symbol 'T' on tape indicating that the checked parentheses expression was correct: for each left parenthesis there was a matching right parenthesis.
Experiment with this program. Enter a new parentheses expression. Click the left button in the desired square to place the yellow cursor in this square and type a parenthesis. Don't forget to place symbols **A** on both sides of the expression. Click the right button in square 1 to position the green cursor in this square. If the yellow cursor is also in this square the square will become blue.
To put system in initial state '0' (*symbol_uttered='0'*) go to block **AM** and click on **teacher**. Position yellow cursor in the square on the right from the name *utter_symbol* and enter '0' in this square. Note that if you are in **memory** mode you cannot enter the symbol. Once the symbol is entered press **Init** button or **F2** key. The initial state is set, that is, *symbol_uttered='0'*. Return to **memory** mode and press **Step** button or **F1** key to do another round of computations.

**Note.** Block **AS** must be in **tape** mode indicating that the robot can see the tape. This block will be in **memory** mode in Example 2 where the robot performs mental computations.

### 1.7.5 Teaching the Robot to Do Parentheses Checking

Write down all twelve commands of the parentheses checker and clear LTM of block **AM** by pressing **Clr G** button. In the following experiment you will teach the robot by entering the output parts of commands (you wrote down) in response to the input parts of these commands. The input parts are displayed in the two upper squares of the XY-column of block **AM**. Block **AM** must be in from **teacher** mode, and block **AS** in **tape** mode.
To teach the robot all twelve commands of the parentheses checker it is sufficient to use the following three training examples: **A(A**, **A)A**, and **A()A** . Write the first expression on tape, place the green cursor in square 1, *set utter_symbol='0'* and press **Init** button. Put block **AM** in learn **new** mode and start pushing **Step** button or **F1** key. See how the new commands (associations) are recorded in LTM of block **AM**.
Repeat the same teaching experiment with other two training examples. If you did everything correctly, all twelve commands are now in LTM. The robot can now perform parentheses checker algorithm with any parentheses expression.

### 1.7.6 Example 2: Performing Mental Computations

Go to **Examples** menu and select Example 2. You can see that both blocks **AM** and **AS** have some programs in their LTM's. In the next section (Example 3) I will explain how the program in block **AS** was created as a result of learning. For now it is sufficient to mention that this program allows this block to simulate the work of external system (W,D) with tape containing up to ten squares, and with external alphabet {'A','(',')','X','T','F'}.
This read/write working memory has limited duration depending on the time constant *tau*. The bigger this time constant the longer the memory lasts. A qualitative theory of this effect was described in Eliashberg (1979). In this study, it will be discussed in Chapter 4. Now I want to show how this "working memory" works. Interestingly enough, block **AS** no longer needs to use its LTM. The effect of read/write working memory is achieved without moving symbols, by simply changing the levels of "residual excitation" of the already stored associations.
Block **AM** has two programs. The program in locations 0-11 is the parentheses checker used in the previous sections. The program in locations 14-26 is a tape scanner. This program starts with the state '3' (*symbol_uttered='3'*). Block **AM** is now in **memory** mode, so the program will run. Block **AS** is in **tape** mode, and the robot can see the tape.
Start pressing **Step** button or **F1** key and see how the robot scans the tape. It first scans to the right, reaches the right boundary, and goes back. Once it reaches the left boundary, it moves to square 1, and changes its "state of mind" to '0'. This transfers control to the parentheses checker (the association in location 0).
At this moment close the robot's eye by clicking left button on the word **memory** in block **AS**. Make sure this word became red. Continue pressing **Step** button or **F1** key and see how the robot performs mental computations.
**Note.** Block **AS** is in learn **none** mode. It could be in learn **all** mode. In this case it would keep recording its XY-sequence. The performance would not change. That is, the effect of read/write working memory is combined with the ability to remember everything.

- *How does block AS work?*

To make it more interesting, I leave it to the reader to figure it out. (Hint. The most recently used association has the highest level of "residual excitation ", e[i], among all competing associations.)

### 1.7.7 Teaching Block AS to Simulate External System (W,D)

Go to **Examples** menu and select Example 3. Block **AM** has two programs in its LTM. The program in locations 0-11 is the program that scans the tape and rewrites it. The program starts in state '8'. In this state (*symbol_uttered='8'*) it rewrites the *symbol_read* and goes to state '9' (*utter_symbol='9'*). In state '9' it moves one step to the right, and goes to state '8'.
Run this program by pressing **Step** button or **F1** key to see what the program is doing. Note that block **AS** is in learn **all** mode so it records the XY-sequence produced by external system (W,D). This is sufficient to learn to simulate this system with the tape, containing no more than ten squares, and with a single external symbol {'A'}.
Prepare the new tape with symbol '(' in squares 0-9. Switch to from **teacher** mode and restore initial state '8'. To do so enter *utter_symbol='8'* and press **Init** button or **F2** key. The *symbol_uttered* is now equal to '8'. Switch back to from **memory** mode, and run the program as before. Block **AS** is now trained to simulate the work of (W,D) with external symbols {'A','('}. Repeat this experiment for the remaining external symbols {')','X','T','F'}. Block **AS** now has the state of LTM identical to that used in Example 1.
Switch block **AS** to learn **none** mode. Write a parentheses expression in squares 0-9 of tape. Don't forget to enter delimiters 'A' on both sides of the expression. The whole expression including delimiters must fit in squares 0-9, because you have taught block **AS** to simulate the tape only for these squares. Put the green cursor representing the scanned square in square 1 by clicking the right button in this square.
To run the test program in locations 14-25 (this program is the same parentheses checker as before) you need to set initial state '0'. To set this state switch to **teacher** mode, enter *utter_symbol='0'*, and press **Init** button or **F2** key. Switch back to **memory** mode and run the program by pressing **Step** button or **F1** key. You are repeating Example 2, but now you trained block **AS** yourself.
**Note.** You could train and test block **AS** in **teacher** mode by doing what was done by the discussed programs. Or you can write your own training and testing programs in block **AM** and run them in **memory** mode.

### 1.7.8 Switching between "External World" and "Imaginary World"

Select Example 4 in **Examples** menu. Block **AM** now has four output channels. The fourth output channel controls closing and opening the robot's eye. ('1' closes the eye and '0' opens it). Run this example by pressing **Step** button or

**F1** key. You will see that block **AS** switches to **memory** mode once scanning is done and parentheses checking starts. From this moment on the robot performs mental computations.

The interesting thing about this example is that the robot itself decides where to switch from "external world" to "imaginary world." This switching doesn't affect the robot's performance, because its mental imagery predicts the results of the robot's actions in the "external world." To make this effect more obvious, select Example 5 and run it. The robot switches between "external world" and "imaginary world" several times while performing computations.

## 1.8 Basic Questions

The cognitive model **(W,D,B)** implemented by the program EROBOT offers some simplified but nontrivial answers to the following set basic questions associated with the brain.

1. *What is the nature of computational universality of the human brain?*
   Computational universality of the human brain is a result of interaction of two associative learning systems: a "processor" system, **AM**, responsible for motor control, and a "memory" system, **AS** responsible for working memory and mental imagery. I intentionally used the terms "processor" and "memory" to emphasize a similarity with the general "processor-memory" architecture of a traditional universal programmable computer.
   The big difference is that, in the discussed model both systems AM and AS are arranged on a similar principle (primitive E-machines). Both "processor" and "memory" are learning systems that create their software (firmware) almost completely in the course of learning. Another big difference is that the "processor" is switching back and forth between "real" world (which serves as external "memory") and "imaginary world" which serves as internal "memory."
   **Remark.** If we identify our "I" with "processor" part, we can say that our "I" is wondering in an infinite loop switching between real and imaginary worlds. The notion of this infinite loop allows our theory to get rid of the notion of "homunculus." The need for a "homunculus" is a result of the lack of universality. If a model of **B(0)** can learn, in principle, to do anything, then a cognitive theory associated with such model doesn't need "homunculus."

2. *How can a person using an external memory device learn to perform, in principle, any algorithm?*
   It is sufficient to memorize SM→M and MS→S associations produced by performing several examples of computations. In the universal learning architecture **(AM,AS)** these associations can serve as software that allows this system to perform the demonstrated algorithm with any input data.
   **Note.** To get the effect of universal programming via associative learning we needed to dedicate a "free" motor channel for the representation of internal states of the finite-state part of simulated Turing machine. In EROBOT this was achieved via "speech" motor channel. This metaphor sheds light on one important role of language. A brain without language cannot achieve the highest level of computing power. Another interesting implication of the EROBOT metaphor is that any sufficiently expressive "free" motor channel can serve as language channel. (Some possibilities of the metaphor the brain as an E-machine with respect to the problem of natural language were discussed in Eliashberg, 1979).

3. *Why does a person performing computations with the use of an external memory device learn to perform similar mental computations?*
   The experiment discussed in Section 1.7.7 provides an answer to this question. While "processor" **AM** interacts with external system **(W,D)** the "memory" system **AS** learns the MS→S associations allowing it to simulate the external system **(W,D)**. Once **AS** is trained, EROBOT can perform mental computations by switching to the "imaginary" system **(W,D)**.

4. *What is the simplest universal learning algorithm consistent with questions 1-3?*
   The simplest universal learning algorithm consistent with questions 1-3 is memorizing all XY-experience. As EROBOT shows, this "dumb" algorithm is not too bad when combined with associative memory in which data is addressed in parallel. The time of **DECODING** doesn't depend on the size of memory (parameter n2 in the neural network of Figure 1.6). The time of **CHOICE** and **ENCODING** is not worse than *log(n2)*. As mentioned in Section 1.5.1, there are many ways to make this algorithm less "memory hungry" and more efficient.

**Note**. A "smart" learning algorithm (such as, for example, backpropagation) is not universal. It optimizes performance in a given context and throws away information needed in a large number of other contexts. I argue that a learning algorithm of this type cannot be employed by the human brain.

5. *What is working memory and mental imagery?*
   The effect of a read/write symbolic working memory is achieved in EROBOT via "dynamical" E-state without moving symbols in "symbolic" LTM. That is, working memory is not a read/write memory buffer. It is one of many effects associated with dynamic reconfiguration of data stored in LTM. The metaphor the brain as an E-machine suggests that such a dynamic reconfiguration is associated with the postulated phenomenological E-states. Once the effect of working memory is produced, system **AS** can work as an "imaginary" external system **(W,D)**. This explains the nature and the functional importance of mental imagery. (There is nothing "imaginary" about mental imagery.)

6. *What is the simplest neural implementation of a model of B(0) consistent with all the above questions?*
   The neural network of Figure 1.6 gives some answer to this question. A connection between phenomenological E-states and the statistical conformational dynamics of ensembles of protein molecules (such as ion channels) is discussed in Eliashberg (1990a, 2003).

# 1.9 Whither from Here

There are several possibilities for the development of the concept of a universal learning neurocomputer illustrated by EROBOT (Eliashberg, 1979, 1989, 1990). These possibilities include:

**1. Introduction of E-states in the Motor Control System (AM)**
This enhancement produces several critically important effects mentioned in Section 1.5.

- Effect of context-dependent dynamic reconfiguration. The same primitive E-machine can be transformed into a combinatorial number of different machines by changing its E-states. No reprogramming is needed!

- Recognition of sequences and effect of temporal associations.

- Effect of "waiting" associations and simulation of stack (with limited depth). This leads to the possibility of calling (and returning from) "subroutines."

- Effect of imitation and "afferent synthesis". A sensory image of a sequence of reactions "pre-activates" (pre-tunes) this sequence. This effect allows the synthesis of complex motor reactions by presenting their sensory images. One can start with "bubbling" and create complex motor sequences. This explains how complex reactions can be learned without the teacher's acting directly on the learner's motor centers (as is done in EROBOT).

**2. Introduction of "centers of emotion" and activating system**
Besides SM→M and MS→S associations employed in EROBOT, the human brain forms associations of other modalities, the most important of which is "emotional" modality. We remember and recognize our emotional states. This means that our brain has some recognizable (encoded) signals that carry information about these states. I use letter "H" to denote "emotional" modality. ("H" stands for "Hedonic". The letter "E" is already used in the name "E-machine".)
Let us assume that besides associative learning systems **AM** and **AS** responsible, respectively, for motor control, and mental imagery, there is also an associative learning system, call it **AH**, responsible for *motivation*. System **AH** forms SH→H associations that serve as "motivational software." (At this general level, I treat "pain" modality as one of "S" modalities.)
This approach can produce much more sophisticated effect of "reward" and "punishment" than is available in the traditional models of reinforcement learning. In the latter models the effect of reinforcement is limited to the modification of SM→M associations.
In the **(AM,AS,AH)** architecture, SH→H associations can interact with SM→M associations via an activating system, the effect of this activation depending on sensory inputs and temporal context. This creates a very complex and interesting situation.

### 3. Introduction of hierarchical structure
Primitive E-machines slightly more complex than Model AF1 described in Section 1.5.6 can be arranged in hierarchical structures. The corresponding complex E-machines can produce various effects of data compression, statistical filtering, and generalization. Some effects of this type were demonstrated in Eliashberg (1979).

### 4. Introduction of "BIOS"
The biggest advantage of the "whole-brain" metaphor "the brain as an E-machine" as compared with traditional "partial" brain modeling metaphors is that the former metaphor has room for arbitrarily complex "brain software." Traditional learning algorithms do create some "neural firmware." This firmware, however, can be compared with the firmware of Programmable Logic Devices (PLD) rather than with the software of a universal programmable computer.

- *How big is brain's "BIOS," and what is in it?*

I believe that this "initial brain's firmware" is the longest part of description of the untrained human brain, **B(0)**. This "BIOS" should include initial "motivational software" (SH→H associations) that determines the direction of learning, and some initial "motor software" (SM→M associations). It may also have some initial software representing initial knowledge about the external world, **W** (represented by MS→S associations).